# Pre-training of Molecular GNNs via Conditional Boltzmann Generator

**Daiki Koge, Naoaki Ono, Shigehiko Kanaya**

Graduate School of Science and Technology, Nara Institute of Science and Technology, Takayama, Ikoma, Nara, Japan.
{koge.daiki.ju9, nono}@is.naist.jp, skanaya@gtc.naist.jp

**Abstract**

Learning representations of molecular structures using deep learning is a fundamental problem in molecular property prediction tasks. Molecules inherently exist in the real world as three-dimensional structures; furthermore, they are not static but in continuous motion in the 3D Euclidean space, forming a potential energy surface. Therefore, it is desirable to generate multiple conformations in advance and extract molecular representations using a 4D-QSAR model that incorporates multiple conformations. However, this approach is impractical for drug and material discovery tasks because of the computational cost of obtaining multiple conformations. To address this issue, we propose a pre-training method for molecular GNNs using an existing dataset of molecular conformations to generate a latent vector universal to multiple conformations from a 2D molecular graph. Our method, called Boltzmann GNN, is formulated by maximizing the conditional marginal likelihood of a conditional generative model for conformations generation. We show that our model has a better prediction performance for molecular properties than existing pre-training methods using molecular graphs and three-dimensional molecular structures.

## 1. Introduction

Learning representations of molecular structures using deep learning is a useful approach in drug and material discovery (Gómez et al. 2018; Stokes et al. 2020; Zhou et al. 2018). In particular, for the task of molecular property prediction, Graph Neural Networks (GNNs) have been successful (Glimer et al. 2018; Duvenaud et al. 2015; Fuchs et al. 2020). Various models of these architectures have been studied, depending on the prediction task. For example, a Graph Field Network (GFN) can predict the potential energy of a molecule from the coordinates of its atomic nucleus (Schütt et al. 2017; Schütt et al. 2018). Although GNNs and GFNs treat molecules as stationary objects, to accurately predict biological or physico-chemical properties, we should use their conformation ensembles. This is because molecules are not static but are in continuous motion in 3D Euclidean space, forming a potential energy surface (PES) (Schlegel et al. 2003; Hawkins et al. 2017). Molecular chemical properties are a function of the set of conformations (conformation ensemble) accessible at a finite temperature (Kuhn et al. 2016).

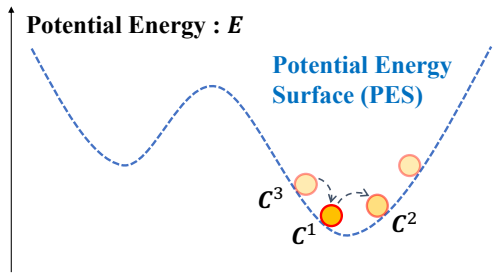

Figure 1 : Illustration of conformations on potential energy surface. The coordinate geometry $C^i$ on the PES is called conformation.

Figure 1 shows the relationship between the PES and conformations.

Recent studies (Zankov et al. 2021; Weinreich et al. 2021; Axelrod et al. 2023) used molecular dynamics (MD) simulations to generate conformation ensembles and used the conformation ensemble as an input to a DNN to predict molecular properties. Here, we refer to these models as 4D-QSARs. These approaches make sense from a physical perspective. However, the use of classical molecular dynamics simulations to explicitly compute a conformation ensemble before predicting its properties is computationally intractable for many real-world applications.

In this study, we propose a pre-training method for GNNs using an existing dataset of conformation ensembles as a surrogate model for 4D-QSARs. From the perspective of statistical physics, a conformation $C$ can be treated as a random quantity sampled from the Boltzmann distribution $p^*(C) \propto \exp(-E(C))$, where $E(C)$ is the potential energy of $C$. If we can obtain a conformation ensemble on the PES as observed samples that follow the $p^*(C)$, we can estimate a universal latent vector for multiple conformations using a conditional generative model.

### 1.1 Related Works

Pre-training methods for Molecular GNNs using molecular conformations have been proposed to obtain a better prediction performance for molecular properties. GraphMVP (Liu et al. 2021) proposes a generative and a contrastive learning task for maximizing between molecular 2D topologies and the 3D conformations. 3D Infomax (Stärk et al. 2022) proposes a knowledge distillation method from a GFN using molecular 3D geometries to GNNs using molecular graphs. These methods use information from multiple conformations but do not explicitly incorporate information on the Boltzmann distribution.

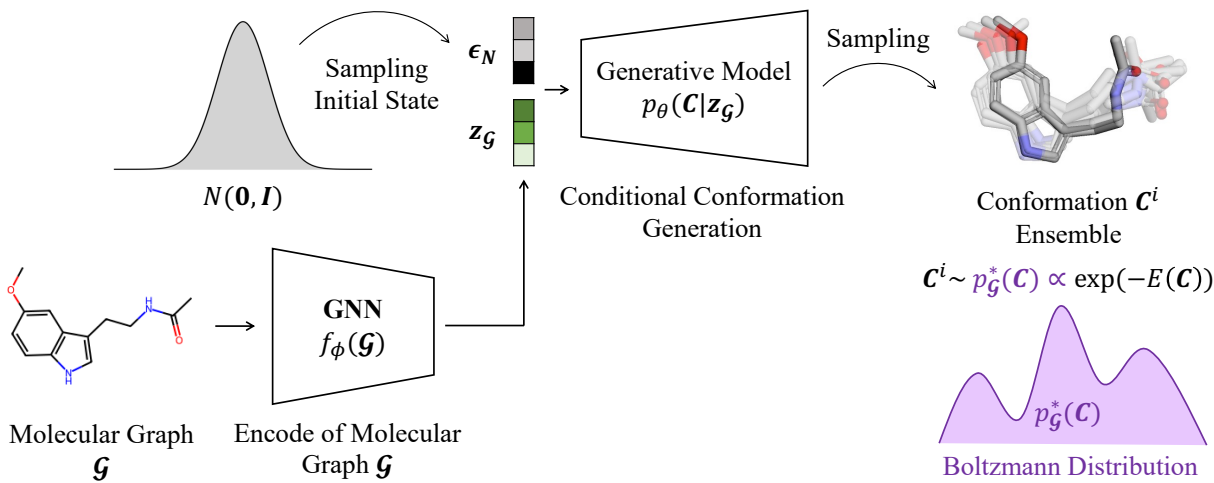

Figure 2 : Scheme of our pre-training method for molecular GNNs. Our pre-training method generates a conditional variable $z_\mathcal{G}$ from molecular graph $\mathcal{G}$ with encoder (GNNs) for $p_\theta(C|z_\mathcal{G})$. This conditional model $p_\theta(C|z_\mathcal{G})$ generates a molecular conformation ensemble using the latent vector $z_\mathcal{G}$.

We introduce a pre-training method for molecular GNNs to incorporate information on the Boltzmann distribution.

## 2. Methods

### 2.1 Preliminaries

**3D conformation of molecule.** For geometry, each atom $a_i$ in molecule $\mathcal{M}$ is embedded by a coordinate vector $c_i \in \mathbb{R}^3$ into 3D space, and the full set of positions (conformation) can be represented as a matrix $C = [c_1, c_2, ... c_n] \in \mathbb{R}^{n \times 3}$.

**2D molecular graph.** A molecular graph is denoted as $\mathcal{G} = (V, E)$, where $V = \{a_i\}_{i=1}^n$ is the set of vertices representing atoms and $E = \{b_{ij} \mid (i,j) \subseteq |V| \times |V|\}$ is the set of edges representing the inter-atomic bonds.

### 2.2 Motivation

Let $p_\mathcal{G}^*(C)$ be the Boltzmann distribution of the conformation ensemble $(C^1, C^2, ... C^N)$ for a molecular graph $\mathcal{G}$. Our aim is to obtain the latent vector $z_\mathcal{G}$ of the molecular graph $\mathcal{G}$ as the conditional variable for a conditional generative model $p_\theta(C|z_\mathcal{G})$ to approximate the $p_\mathcal{G}^*(C)$. This problem is to obtain $z_\mathcal{G}^* \in \mathbb{R}^D$ such that,

$$z_\mathcal{G}^* = \underset{z_\mathcal{G} \in \mathbb{R}^D}{\operatorname{argmin}} D_{\mathrm{KL}}[p_\mathcal{G}^*(C) || p_\theta(C|z_\mathcal{G})], \quad (1)$$

where $\theta$ is a set of parameters for conditional generative model $p_\theta(C|z_\mathcal{G})$. If $N$ is large, $D_{\mathrm{KL}}[\cdot]$ of Eq. 1 can be rewritten as follows:

$$\widehat{z_\mathcal{G}} = \underset{z_\mathcal{G} \in \mathbb{R}^D}{\operatorname{argmin}} \left[ \frac{1}{N}\left(-\sum_{i=1}^N \log(p_\theta(C^i|z_\mathcal{G}))\right) + H(p_\mathcal{G}^*(C)) \right]. \quad (2)$$

$H(p_\mathcal{G}^*(C))$ is the entropy of $p_\mathcal{G}^*(C)$ and constant term. Therefore, Eq. 3 represents the maximum likelihood estimator (MLE). This maximum likelihood estimate $\widehat{z_\mathcal{G}}$ includes information on the conformation ensemble $(C^1, C^2, ... C^N)$ on the PES. In practice, the latent vector encoded by a $f_\phi(\mathcal{G})$ (GNN) is denoted as $z_\mathcal{G}$, and we estimate the parameter $\hat{\phi}$ of the $f_\phi(\mathcal{G})$ using MLE. Therefore, we can obtain the following objective:

$$\hat{\phi} = \underset{\phi \in \mathbb{R}^d}{\operatorname{argmin}} \left[ \frac{1}{N}\left(-\sum_{i=1}^N \log(p_\theta(C^i|f_\phi(\mathcal{G})))\right) \right]. \quad (3)$$

We use this MLE as an objective function for training GNNs. Our goal is to improve the prediction performance of molecular properties for small datasets using this training method for the pre-training of GNNs.

Figure 2 shows a schematic illustration of our solution, which constructs a conditional Boltzmann generator that approximates the Boltzmann distribution using a hierarchical model of a GNN and conditional generative model.

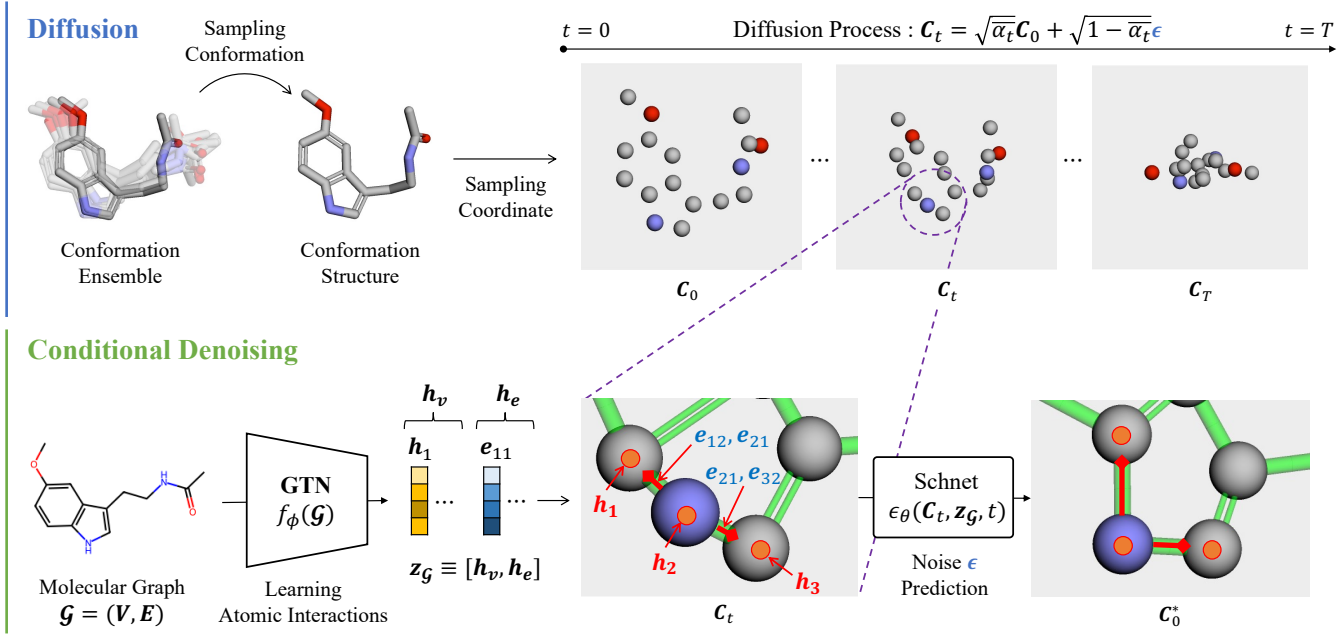

Figure 3 : Illustration of our pre-training method. First, the diffusion process adds a noise $\epsilon$ to the coordinates of a conformation $C_0$. GTN encodes a molecular graph $\mathcal{G}$ to the latent vector $z_\mathcal{G}$, and Schnet uses it to predict the noise $\epsilon$. $C_0^*$ indicates the original conformation before noise is added.

### 2.3 Model Architecture

**Conditional generative model.** Conditional generative model $p_\theta(C|z_\mathcal{G})$ needs to satisfy SE(3)-invariant likelihood. Furthermore, it should be possible to generate multimodal distributions, such as the Boltzmann distribution. To satisfy these requirements, we use a geometric diffusion model (Geodiff) (Xu et al. 2022).

Geodiff is a conditional generative model using denoising diffusion probabilistic model (DDPM) (Ho et al. 2020). The training process of Geodiff is to predict Gaussian noise $\varepsilon$ from a molecular graph $\mathcal{G}$ and a noisy conformation $C_t$ that contains a Gaussian noise $\varepsilon_t \in \mathbb{R}^{n \times 3} \sim N(0, I)$. This noisy conformation is obtained by a discrete Markov chain, called the diffusion process, using the following equation:

$$C_t = \sqrt{1-\beta_t} C_{t-1} + \sqrt{\beta_t} \varepsilon_t, \quad (4)$$

where $t$ represents time, and $C_0 = C$. By increasing $\beta_t$ from 0 to 1 as $t$ increases, $C_0$ is converted to a random noise vector. Let $\alpha_t \equiv 1 - \beta_t$ and $\overline{\alpha_t} \equiv \prod_{s=1}^{t} \alpha_s$, we get a sample $C_t$ with noise $\epsilon \in \mathbb{R}^{n \times 3} \sim N(0, I)$ from following equation:

$$C_t = \sqrt{\overline{\alpha_t}} C_0 + \sqrt{1-\overline{\alpha_t}} \epsilon. \quad (5)$$

The objective of Geodiff is to minimize the following equation as a variational upper bound on the negative log marginal likelihood $-\log p_\theta(C|\mathcal{G})$:

$$\mathbb{E}_{t \sim \text{Uniform}(\{1,T\})} [\|\epsilon - \epsilon_\theta(C_t, \mathcal{G}, t)\|^2]. \quad (6)$$

$\epsilon_\theta$ is Schnet (Schütt et al. 2018). In Geodiff, although the input $\mathcal{G}$ of Schnet is fixed, we change it to a latent vector $z_\mathcal{G}$ with $\mathcal{G}$ encoded by GNNs ($f_\phi : \mathcal{G} \to z_\mathcal{G}$). Thus, we change Eq. 6 to follows:

$$\mathbb{E}_{t \sim \text{Uniform}(\{1,T\})} [\|\epsilon - \epsilon_\theta(C_t, z_\mathcal{G}, t)\|^2]. \quad (7)$$

**GNN model for encoding molecular graphs.** We use Graph transformer network (GTN) (Dwivedi et al. 2020) for encoding molecular graphs into their latent vectors $z_\mathcal{G}$. GTN uses self-attention (Vaswani et al. 2017) and Laplacian encoding to embed atomic interactions in a molecular graph $\mathcal{G} = (V, E)$ into latent atomic vectors $h_v = \{h_i\}_{i=1}^{n}$ and latent edge vectors $h_e = \{e_{ij} \mid (i,j) \subseteq |V| \times |V|\}$. We define $z_\mathcal{G}$ as $[h_v, h_e]$. The encoding of GTN ($f_\phi : \mathcal{G} \to z_\mathcal{G}$) is

$$h_v, h_e = f_\phi(\mathcal{G}). \quad (8)$$

Figure 4 shows our method using GTN and Geodiff. Schnet $\epsilon_\theta$ predicts a noise $\epsilon$ added to the conformation $C$, which is called score function. This model estimates the molecular force field (Zaidi et al. 2022). Therefore, $f_\phi(\mathcal{G})$ learns latent vectors $z_\mathcal{G}$ about atomic interactions on the conformation. Latent edge vectors $h_e$ are important for extracting the atomic interactions between atoms.

Table 1: Results for molecular property prediction tasks. For each downstream task, we report the mean squared error (MSE) of 3 seeds with scaffold splitting. The best performance for each task is marked in **bold**. BACE1 and CTSD are biological activity datasets for a target protein from Excape-DB. We wrote the sample size next to the dataset name.

| Pre-training method | Small Datasets (Sample size) | | | | |
|---|---|---|---|---|---|
| | Solubility (1.1k) | Malaria (10k) | Lipophilicity (4.2k) | BACE1 (3.6k) | CTSD (1.1k) |
| GraphCL | 1.2189 | 1.2152 | 0.5708 | 0.8409 | 0.8066 |
| AttrMask | 1.3396 | 1.2522 | 0.5437 | 0.7672 | 0.7268 |
| 3D Infomax | 1.2276 | 1.2263 | 0.5389 | 0.7882 | 0.9327 |
| GraphMVP | 1.1719 | 1.1619 | **0.5088** | 0.8058 | 0.7689 |
| Boltzmann GNN | **0.8649** | **1.1586** | 0.6346 | **0.5984** | **0.7257** |

## 2.4 Loss Function

The loss function is the mean of the evidence upper bound on the $-\log p_\theta(C^i | f_\phi(\mathcal{G}))$:

$$\frac{1}{N}\left(\sum_{i=1}^{N} \mathbb{E}_{t \sim \text{Uniform}(\{1,T\})} \left[\left\| \epsilon - \epsilon_\theta\left(C_t^i, f_\phi(\mathcal{G}), t\right)\right\|^2\right]\right), \quad (9)$$

where $C_t^i$ is the conformation corresponding to the molecular graph $\mathcal{G}$, and is the noisy conformation obtained from time $t$ of the diffusion process. In practice, we compute the above loss Eq. 9 for the various molecular graphs and conformation ensembles in a dataset. We minimize the following expectation $\mathcal{L}(\theta, \phi)$ as in Geodiff:

$$\mathbb{E}_{(c^i, \mathcal{G}) \sim \pi(C, \mathcal{G}), t \sim \text{Uniform}(\{1,T\})} \left[\left\| \epsilon - \epsilon_\theta\left(C_t^i, f_\phi(\mathcal{G}), t\right)\right\|^2\right].$$

$\pi(C, \mathcal{G})$ is a joint distribution of molecular graphs and conformations obtained from a dataset. We can optimize $\theta$ and $\phi$ with stochastic gradient descent for $\mathcal{L}(\theta, \phi)$. We call GNNs using this objective Boltzmann GNN.

## 3. Experiment and Results

We empirically evaluate our model with transfer learning tasks for small datasets. We compare the performance of the Boltzmann GNN with existing pre-training methods.

### 3.1 Setup

**Datasets.** For the pre-training datasets, we take 60k molecules from GEOM (Axelrod et al. 2022). We took 5 conformers for each molecule. For downstream tasks, we obtained datasets of biological activity to target proteins obtained from Excape-DB (Sun et al. 2017) and datasets of physico-chemical properties such as solubility and lipophilicity from MoleculeNet (Wu et al. 2018). Finally, we set five regression tasks.

**Baselines.** For 2D graph-based pre-training methods, we chose well-acknowledged SSL methods: GraphCL (You et al. 2020) and AttrMask (Hu et al. 2019). For 3D structure informed pre-training methods, we chose recent proposed SSL methods: GraphMVP (Liu et al. 2021) and 3D Infomax (Stärk et al. 2022). We used Graph Isomorphism Network (GIN) (Xu et al. 2018) and Schnet as baseline models.

**Pre training.**
We trained our model and baseline models for 500 epochs and determined the best model for each on validation samples not used for training.

### 3.2 Results

We summarized in Table 1 the mean squared error for each model in each dataset. Boltzmann GNN achieved state-of-the-art performance for four of the five tasks. Furthermore, our model performed better for datasets with small sample sizes such as Solubility and BACE1.

## 4. Conclusion and Future work

In this study, we proposed a novel pre-training method for molecular GNNs via conditional Boltzmann generator. We integrated the geometric diffusion model and the graph transformer to infer the latent vector of the Boltzmann distribution. Our pre-training method explicitly incorporated information on Boltzmann distribution, which improved prediction performance for downstream tasks such as molecular properties.

These results support the effectiveness of the pre-training method with conditional Boltzmann generation, and we will continue to explore further in this direction.

## 5. Acknowledgement

This work was supported by JSPS KAKENHI (Grant number 22J11040).